\documentclass[runningheads]{llncs}

\usepackage{caption} 
\usepackage{booktabs}
\usepackage{multirow}

\usepackage{nicefrac}

\usepackage{algorithm}
\usepackage[noend]{algpseudocode}
\usepackage{textcomp}

\usepackage{xcolor}

\usepackage{subfigure}

\usepackage{hyperref}
\usepackage{comment}

\usepackage{amssymb,amsmath}
\usepackage{mathtools}
\newtheorem{fact}{Fact}

\usepackage{stmaryrd}

\usepackage{xfp}

\usepackage{tikz}
\usepackage{tikzscale}
\usepackage{xspace}
\usepackage{pgfplots}
\usepgfplotslibrary{groupplots}
\pgfplotsset{compat=1.16}
\usepackage{xparse}
\NewDocumentCommand{\Log}{o}{%
  \IfNoValueTF{#1}{}{{}^{#1}\!}\log}%

\newcommand{\vect}[1]{\mathbf{#1}}

\newcommand{\tikzxmark}{%
\tikz[scale=0.23] {
    \draw[line width=0.7,line cap=round] (0,0) to [bend left=6] (1,1);
    \draw[line width=0.7,line cap=round] (0.2,0.95) to [bend right=3] (0.8,0.05);
}}

\usepackage[switch]{lineno}

\newcommand{\eg}{\emph{e.g.}}

\usepackage{breakurl}

\title{BioSequence2Vec: Efficient Embedding Generation For Biological Sequences}

\author{Sarwan Ali\inst{1}, Usama Sardar\inst{2}, Murray Patterson\inst{1,*}, Imdad Ullah Khan\inst{2,*}}%
\institute{
Georgia State University, Atlanta, GA, USA \\
\email{sali85@student.gsu.edu, mpatterson30@gsu.edu} \and
Lahore University of Management Sciences, Lahore, Pakistan\\
\email{usamasardar2022@gmail.com, imdad.khan@lums.edu.pk}\\
* Corresponding authors, Joint Last Authors}

\begin{document}

\maketitle

\begin{abstract}
Representation learning is an important step in the machine learning pipeline. Given the current biological sequencing data volume, learning an explicit representation is prohibitive due to the
dimensionality of the resulting feature vectors.  Kernel-based methods, \eg, SVM, are a proven efficient and useful alternative for several machine learning (ML) tasks such as sequence classification.
Three challenges with kernel methods are (i) the computation time, (ii) the memory usage (storing an $n\times n$ matrix), and (iii) the usage of kernel matrices limited to kernel-based ML methods (difficult to generalize on non-kernel classifiers).  While (i) can be solved using \textit{approximate} methods, challenge (ii) remains for typical kernel methods.  Similarly, although non-kernel-based ML methods can
be applied to kernel matrices by extracting principal components (kernel PCA), it may result in information loss, while being computationally expensive. In this paper, we propose a general-purpose representation learning approach that embodies kernel methods' qualities while avoiding computation, memory, and generalizability challenges.  This involves computing a low-dimensional embedding of each sequence, using random projections of its $k$-mer frequency vectors, significantly reducing the computation needed to compute the dot product and the memory needed to store the resulting representation. Our proposed fast and alignment-free embedding method can be used as input to any distance (\eg, $k$ nearest neighbors) and non-distance (\eg, decision tree) based ML method for classification and clustering tasks.  Using different forms of biological sequences as input, we
perform a variety of real-world classification tasks, such as SARS-CoV-2 lineage and gene family classification, outperforming several state-of-the-art embedding and kernel methods in predictive performance.
\end{abstract}


\section{Introduction}
The rate at which biological sequence data is being generated and
stored currently outpaces Moore's law growth, even under fairly
conservative estimates~\cite{stephens-2015-genomical}.  In the past
decade, the amount of biological sequence data has already reached a
level that automated --- machine learning (ML) and deep learning (DL)
--- algorithms are able to learn from and perform analysis.  A
notable example is the AlphaFold framework~\cite{AlphaFold2021} for
structure prediction from a protein sequence.  Another example is the
Pangolin tool~\cite{otoole-2021-pangotool} for predicting lineage,
\eg, B.1.1.7~\cite{rambaut-2020-pangonomenclature}, from an assembled
SARS-CoV-2 genomic sequence.  The number of SARS-CoV-2 genomic
sequences which are publicly available on databases such as
GISAID\footnote{\url{https://www.gisaid.org/}} is more than 15 million and counting.

While approaches such as AlphaFold and Pangolin have outperformed the
previous state-of-the-art by a significant margin, or can scale to
orders of magnitude more in number of sequences, these learning
approaches can be and will need to be optimized, as the amount of
data grows to strain even these approaches.
Since the majority of ML models deal with fixed-length numerical
vectors, biological sequences cannot be used directly as input, making efficient embedding design an important step in such ML-based pipelines~\cite{hu2022learning,chourasia2022clustering}.

Sequence alignment is another important factor to be considered while
performing sequence analysis.  Although embeddings that require the
sequences to be aligned, such as one-hot encoding
(OHE)~\cite{kuzmin2020machine}, are proven to be efficient in terms of
predictive performance, researchers are interested in exploring
alignment-free methods to avoid the expensive
multiple sequence alignment operations as a preprocessing
step~\cite{ali2022evaluating,ali2023characterizing,chourasia2022informative,chowdhury2017review,tayebi2021robust}. Most alignment-free methods compute
some form of a sketch of a sequence from short substrings, such as
$k$-mers to generate a spectrum~\cite{ali2021k}. Although the existing alignment-free
embedding methods yield promising predictive results; they produce vectors of high
dimensionality, especially for very long sequences.
An alternative to the traditional \textit{feature engineering} methods
is using deep learning (DL) models~\cite{ali2022spike2signal}. However, DL methods have not
seen much success in the classification of tabular
datasets~\cite{borisov2021deep,ali2022benchmarking}.

Using a kernel (Gram) matrix for sequence classification and
kernel-based ML classifiers, such as SVM, shows promising
results~\cite{farhan2017efficient,ali2022efficient_tcbb}. Kernel-based methods outperform feature engineering-based
methods~\cite{ali2021k}. These methods work by computing kernel
(similarity) values between pairs of sequences based on the number of
matches and mismatches between their
$k$-mers~\cite{ali2022efficient_tcbb}. The resultant kernel
matrix can then be used to classify the sequences using SVM.
However, there are serious challenges to the scalability of kernel-based methods to large datasets:
\begin{itemize}
\item Evaluating the kernel value between a pair of sequences takes time proportional to $|\Sigma|^k$, where $\Sigma$ is the alphabet of sequences;
\item Storing the $n \times n$ kernel matrix is memory intensive
  when $n$ is very large; and
\item Kernel matrices are limited to kernel-based machine learning
  models (such as SVM) downstream.
\end{itemize}
The first challenge of kernel evaluation can be
overcome with the so-called \textit{kernel trick} and approximating kernel values with quality guarantees~\cite{farhan2017efficient}.  To use
more general classifiers like decision trees, one can compute principal components of the kernel matrix using kernel PCA~\cite{hoffmann2007kernel}, which
can act as the embedding representation to tackle the third challenge. However, this process 
results in a loss of information and is computationally
expensive.  In general, the second challenge of the need to store an $n
\times n$ kernel matrix in memory remains unaddressed.

In this paper, we propose a random projection-based sequence representation called  BioSequence2Vec, which has the qualities of kernel methods in terms of efficiently computing  pairwise similarity between sequences (overcoming the first challenge) while also addressing the memory overhead. Given a (biological) sequence as input, the BioSequence2Vec embedding projects frequency vectors of all $k$-mers in a sequence in ``random directions'' and uses these projections to represent the sequence. BioSequence2Vec computes the projections in one linear scan of the sequence (rather than explicitly computing the frequency of each of the $|\Sigma|^k$ $k$-mers in the sequence). Since our method computes the representation of a sequence in linear time (linear in the length of the sequence), it easily scales to a large number of sequences. The generated embeddings are low-dimensional (user-controlled), hence BioSequence2Vec overcomes the memory usage problem. The Euclidean (and cosine) similarity between a pair of embeddings is closely related to the kernel similarity of the pair, hence our method incorporates the benefits of kernel-based methods. BioSequence2Vec is efficient, does not require sequences to be aligned, and the embeddings can be used as input to any distance (\eg, $k$ nearest neighbors) and non-distance (\eg, decision tree) based ML methods for the supervised tasks, solving the third problem.

In summary, our contributions are the following:
\begin{enumerate}
\item We propose a fast, alignment-free, and efficient embedding
  method, BioSequence2Vec, which quickly computes a low
  dimensional numerical embedding for biological sequences. It has the
  quality of kernel-based methods in terms of computing pair-wise
  similarity values between sequences while also addressing the
  memory usage issue of kernel methods --- allowing it to scale to
  many more sequences.
\item We show that the proposed method can be generalized on
  different types of real-world biological sequences. It outperforms
  both alignment-based and alignment-free SOTA methods for
  predictive performance on different datasets.
\item Our method eliminates the expensive multiple sequence alignment step from the classification pipeline, hence making it a fast and scalable approach.
\end{enumerate}

The rest of the paper is organized as follows: The literature for biological sequence analysis is given in
Section~\ref{sec_related_work}. Section~\ref{sec_ourSol} outlines the details of the proposed model. The description of the dataset and experimental setup are given in
Section~\ref{sec_datasetDes}. The empirical results are provided in Section~\ref{sec_results}.
Section~\ref{sec_conclusion} concludes the paper and discusses future prospects.

\section{Related Work}\label{sec_related_work}
Designing numerical embeddings is an important step in the ML pipeline for supervised analysis~\cite{ali2022efficient,ullah2020effect}.
The feature engineering-based methods, such as Spike2Vec~\cite{ali2021spike2vec} and PWM2Vec~\cite{ali2022pwm2vec}, which are based on the idea of using $k$-mers achieve reasonable predictive performance. However, they still face the problem of \textit{curse of dimensionality}. As we increase the value of $k$, the spectrum (frequency count vector) becomes sparse (contains small $k$-mers counts). Hence the likelihood of observing a specific $k$-mer again decreases.
To solve this sparse vector problem, authors in~\cite{ghandi2014robust} propose the idea of using gapped/spaced $k$-mer. 
The use of $k$-mers counts for phylogenetic applications was first explored in~\cite{Blaisdell1986AMeasureOfSimilarity}, which constructed accurate phylogenetic trees from coding and non-coding nDNA sequences. Although phylogenetic-based methods are useful for sequence analysis, they are computationally expensive, hence cannot be scaled on bigger datasets.

Computing the pair-wise similarity between sequences by computing kernel/gram matrix is a well-studied problem in ML domain~\cite{ali2021k}. Since computing the pair-wise similarities could be expensive to compute, authors in~\cite{farhan2017efficient} proposed an approximate method to improve the kernel computation time by computing the dot product between the spectrum of two sequences. The resultant kernel matrix can be used as input for kernel classifiers such as SVM or non-kernel classifiers~\cite{ali2021k} using kernel PCA~\cite{hoffmann2007kernel} for classification purposes. 

Authors in~\cite{shen2018wasserstein} propose a neural network-based model to
extract the features using the Wasserstein distance. 
A ResNet model for the purpose of classification is proposed in~\cite{wang2017time}.
However, DL methods, in general, do not show promising results when applied to tabular data~\cite{shwartz2022tabular}.
Using pre-trained models is also explored in the literature for biological sequence analysis~\cite{heinzinger2019modeling,brandes2022proteinbert}. 
However, since those models are usually trained on a specific type of biological sequence, they cannot easily be generalized on different types of data.

\section{Proposed Approach}\label{sec_ourSol}  
In this section, we describe the details of our sequence
representation, BioSequence2Vec. We also analyze the space and time
complexity of computing the representations. As outlined above,
sequences generally have varying lengths, and even when the lengths
are the same, the sequences may not be aligned. Thus, they cannot be
treated as vectors. Though in aligned sequences of equal length, a
one-to-one correspondence between elements is established, treating
them as vectors ignores the order of elements and their contiguity. In
one of the most successful approaches that cater to all of the above
issues, sequences are represented by fixed-dimensional feature
vectors. The feature vectors are the spectra, or counts, of all
$k$-mers appearing in the
sequences~\cite{ali2021spike2vec}.

Suppose we are given a set ${\cal S}$ of sequences (of, \eg,
nucleotides A, C, G, T). For a fixed integer $k$, let $\Sigma^k$ be the
set of all strings of length $k$ made from characters in $\Sigma$ (all
possible $k$-mers) and let $s=|\Sigma|^k$. The spectrum $\Phi_k(X)$ of
a sequence $X\in {\cal S}$ is a $s$-dimensional vector of the counts
of each possible $k$-mer occurring in $X$.  More formally,
\begin{equation}\label{defspectrum}
  \Phi_{k}(X) =  \left( \Phi_{k}(X)[\gamma]\right)_{\gamma \in \Sigma^k} 
=     \left( \sum_{\alpha \in X} I(\alpha,\gamma)\right)_{\gamma \in \Sigma^k},
\end{equation}
where 
\begin{equation}
    I_k (\alpha,\gamma) = 
    \begin{cases}
    1,& \text{if } \alpha = \gamma  \\
    0,              & \text{otherwise}
    \end{cases}
\end{equation}
Note that $\Phi_{k} (X)$ is a $s=\vert \Sigma \vert^k$-dimensional
vector where the coordinate $\gamma \in \Sigma^k$ has a value equal to
the frequency of $\gamma$ in $X$. Since this dimensionality grows
quickly for modest-sized alphabets --- it is exponential in $k$ ---
the space complexity of representing sequences can quickly become
prohibitive.

In kernel-based machine learning, a {\em kernel function} computes a real-valued similarity score for a  pair of feature vectors. The kernel function is typically the inner product of the respective spectra. 
\begin{equation}\label{defkernelValue}
\begin{aligned}
    K(i,j) = K(X_i,X_j) = \langle  \Phi_k (X_i), \Phi_k (X_j) \rangle
    \\
     = \Phi (X_i) \cdot \Phi (X_j)
          = \sum_{\gamma \in \Sigma^k} \Phi_k(X_i) [\gamma] \times \Phi_k(X_j) [\gamma]
\end{aligned}
\end{equation}
The kernel matrix (of pairwise similarity scores) is input to a
standard support vector machine
(SVM)~\cite{cristianini2000introduction}
classifier resulting in excellent classification performance in many
applications~\cite{farhan2017efficient}. Although,
in the so-called {\em kernel trick}, the explicit computation of
feature vectors are avoided, with quadratic space required to store the
kernel matrix, even this approach does not scale to real-world
sequences datasets.
There are three challenges to overcome: (i) Explicit representation is prohibitive due to the dimensions of the feature vectors, (ii) Although explicit computation is avoided using the kernel trick~\cite{farhan2017efficient}, the storage complexity of the kernel matrix is too large, and (iii) Kernel methods do not allow non-kernel-based machine learning methods.
In the following, we propose a representation learning approach,
namely BioSequence2Vec, that encompasses the benefits of the
kernels and allows employing both kernel-based and general purpose
machine learning methods.

\subsection{BioSequence2Vec Representation}

The BioSequence2Vec representation, $\vect{\hat{x}}$ for a sequence
$X$ represents $X$ by the random projections of $\Phi_k(X)$ on the
``discrete approximations'' of random directions. It allows the
application of vector space-based machine learning methods. We show
that the Euclidean distance between a pair of vectors in
BioSequence2Vec representation is closely related to the kernel-based
proximity measure between the corresponding sequences. We use $4$-wise
independent hash functions to compute $\Phi'(\cdot)$.
Note that the definition of our representation
of~\eqref{sketch} is inspired by the work of AMS~\cite{alon1996space}
to estimate the frequency moments of a stream.

\begin{definition}[$4$-wise Independent hash function]
A family ${\cal H}$ of functions of the form $h:\Sigma^k \mapsto
\{-1,1\}$ is called $4$-wise independent, or $4$-universal, if a
randomly chosen $h\in {\cal H}$ has the following properties:
\begin{enumerate} 
\item for any $\alpha\in \Sigma^k$, $h(\alpha)$ is equally likely to
  be $-1$ or $1$.
\item for any distinct $\alpha_i \in \Sigma^k, \text{ and } m_i \in
  \{-1,1\} \;\; (1\leq i \leq 4)$, $$ Pr[h(\alpha_1) = m_1 \wedge
    \ldots \wedge h(\alpha_4) = m_4] = \nicefrac{1}{2^4}$$
\end{enumerate}
\end{definition}

Next, we give a construction of a $4$-wise independent family of hash
functions due to Carter and Wegman~\cite{carter1977universal}

\begin{definition}
Let $p $ be a large prime number. For integers $a_0, a_1, a_2, a_3$,
such that $0\leq a_i \leq p-1$ , and $\alpha\in \Sigma^k$ (represented
as integer base $|\Sigma|$), the hash function $h_{a_0, a_1, a_2,
  a_3}: \Sigma^k \mapsto \{-1,1\}$ is defined
as \begin{equation}\label{eq_h_val} h_{a_0, a_1, a_2, a_3}(\alpha)
  = \begin{cases} -1 & \text{ if } g(\alpha) =0 \\ 1 & \text{ if }
    g(\alpha) =1 \end{cases}
\end{equation}

where  
\begin{equation}\label{eq_h_val_2}
    g(\alpha) =  \big(a_0 + a_1 \alpha + a_2\alpha^2 + a_3\alpha^3 \mod p \big) \mod 2
\end{equation}
\end{definition}

It is well-known that the family ${\cal H} = \{h_{a_0, a_1, a_2, a_3}:
0\leq a_i<p \}$ is $4$-universal. Choosing a random function from this
family amounts to choosing four random coefficients of
polynomial, and the hash value for a $k$-mer $\alpha$ is the value of
the polynomial (with random coefficients) at $\alpha$ modulo the prime
$p$ and modulo $2$.

We use the following property of any randomly chosen function $h$ from ${\cal H}$ that directly follows from the definition.

\begin{fact}\label{factIndependence}
For any distinct $\alpha_1, \alpha_2 \in \Sigma^k$, $E[h(\alpha_2) h(\alpha_2)] = 0$
\end{fact}
The property of $4$-wise independence is used to derive a bound on the variance of the inner product.  Let $t$ be a fixed positive integer (a user-specified quality parameter). For $1\leq i \leq t$, let $h^{(i)}
= h^{(i)}_{a_0,a_1,a_2,a_3}$ be $t$ randomly and independently chosen
functions from ${\cal H}$ (corresponds to choosing $t$ sets of $4$
integers modulo $p$).

The $i$th coordinate of our representation, $\vect{\hat{x}}$ of a sequence $X$ is given by

\begin{equation}\label{sketch}
\vect{\hat{x}}_i  = \dfrac{1}{\sqrt{t}} \sum_{\alpha \in X} h^{(i)}(\alpha) .\end{equation}

In other words, The $i$th coordinate is the projection on the random vector in $\mathbb{R}^{|\Sigma|^k}$, a corner of the $|\Sigma|^k$-dimensional hypercube. More precisely, $\vect{\hat{x}}$ is a $t$-dimensional vector, where the value at the $i$th coordinate is the (normalized) dot-product of $\Phi_k(X)$ with the vector in $\{-1,1\}^{|\Sigma|^k}$ given by $h^{(i)}$. 

Next, we show that the dot-product between the BioSequence2Vec representations $\vect{\hat{x}}$ and $\vect{\hat{y}}$ of a pair of sequences $X$ and $Y$ closely approximates the kernel similarity value given in~\eqref{defkernelValue}. We are going to show that for any pair of sequences $X$ and $Y$,  $\vect{\hat{x}}\cdot \vect{\hat{y}} \simeq \Phi_k (X) \cdot \Phi_k (Y)$. For notational convenience let $\vect{x} = \Phi_k(X)$ and  $\vect{y} = \Phi_k (Y)$, we show that $\vect{\hat{x}}\cdot \vect{\hat{y}} \simeq \vect{x}\cdot \vect{y} $.

\begin{theorem}\label{qualityThm}
For any $0 < \epsilon, \delta < 1$, if  $t\geq \nicefrac{2}{\epsilon^2} \log(\nicefrac{1}{\delta}) $, then 

\begin{enumerate}
\item $E\big[\vect{\hat{x}}\cdot \vect{\hat{y}}\big]  =  \vect{x} \cdot \vect{y}$

\item $Pr\big[ \vert \vect{\hat{x}}\cdot \vect{\hat{y}} - \vect{x} \cdot \vect{y}| \leq \epsilon \|\vect{x}\| \|\vect{y} \| \big] \geq 1-\delta$
\end{enumerate}

\end{theorem}

The proof of 1. will be provided in the full version of the paper.

The proof of 2. follows from a standard application of Hoeffding's inequality. First note that by construction for $1\leq i \leq t$, we have $$-\nicefrac{\|\vect{x}\|}{\sqrt{t}} \leq \vect{\hat{x}}_i \leq  \nicefrac{\|\vect{x}\|}{\sqrt{t}} $$ Similar bounds hold on each coordinate of $\vect{\hat{y}}$. Also note that $\|\vect{x}\| = \| \Phi_k(X) \|$ is the number of $k$-mers in $X$. These bounds implies that $$-\nicefrac{\|\vect{x}\|\|\vect{y}\|}{{t}} \leq \vect{\hat{x}}_i\times \vect{\hat{y}}_i  \leq  \nicefrac{\|\vect{x}\|\|\vect{y}\|}{{t}} $$  
Using these bounds in Hoeffding's inequality, we get that $$Pr\big[ \vert \vect{\hat{x}} \cdot \vect{\hat{y}} -  \vect{x}\cdot \vect{y} \vert \geq \epsilon \|\vect{x}\| \|\vect{y} \| \big] \leq  e^{\nicefrac{- t \epsilon^2}{2}}. $$ Substituting the value of $t$ we get the desired probabilistic guarantee on the quality of our estimate. \qed
Note that the upper bound on the error is very loose, in practice we get far better estimates of the inner product.

\begin{remark}
The runtime of computing $\vect{\hat{x}}$ is $tn_x$, where $n_x$ is the number of characters in $X$. The space complexity of saving $\vect{\hat{x}}$ is $\nicefrac{2}{\epsilon^2}\log(\nicefrac{1}{\delta})$, where both $\epsilon$ and $\delta$ are user-controlled parameters. In the error term, $\|\vect{x}\| = n_x -k+1$, when $\vect{x}  = \Phi_k(X)$.
\end{remark}

Next, we show that the $\ell_2$-distance between any two
vectors, which is usually employed in vector-space machine learning
methods (e.g. $k$-NN classification) is
closely related to their inner product. The inner product of the
BioSequence2Vec representations of two sequences closely approximate the kernel similarity score between two sequences, see Equation~\eqref{defkernelValue}. Thus, BioSequence2Vec achieves the benefits of
kernel-based learning while avoiding the time complexity of kernel
computation and the space complexity of storing the kernel matrix.

Suppose we scale the BioSequence2Vec representation $\vect{\hat{x}}$
of the sequence $X$ by $\|\vect{\hat{x}}\|_2$ ($\ell_2$ norm of
$\vect{\hat{x}}$, to make them unit vectors). Then, by definition, we get the following relation between $\ell_2$-distance and inner
product between $\vect{\hat{x}}$ and $\vect{\hat{y}}$.
\begin{align*}
    d^2 (\vect{\hat{x}},\vect{\hat{y}}) &= \sum_{i=1}^{t} (\vect{\hat{x}}_i - \vect{\hat{y}}_i) = \sum_{i=1}^{t} \vect{\hat{x}}_i^2 + \sum_{i=1}^{t} \vect{\hat{y}}_i^2 - 2 \sum_{i=1}^{t} \vect{\hat{x}}_i \vect{\hat{y}}_i
    \\
   &= 1 + 1 - 2 (\vect{\hat{x}} \cdot \vect{\hat{y}}) = 2 - 2 \; Cos \; \theta_{\hat{x}\hat{y}}
\end{align*}
where $\theta_{uv}$ is the angle between the $\vect{u}$ and $\vect{v}$
in $\mathbb{R}^t$.  Thus, both the ``Euclidean and cosine similarities''
between two BioSequence2Vec vectors are proportional to the ``kernel
similarity'' between the corresponding sequences.

The pseudocode of BioSequence2Vec is given in
Algorithm~\ref{algo_BioSequence2Vec}. 
Our algorithm takes as input a set ${\cal S}$ of biological sequences, integers $k$, $p$, alphabet $\Sigma$, and the number of hash functions $t$. It outputs the embedding $R$, which is the $t$ dimensional fixed-length numerical representation corresponding to set ${\cal S}$ of sequences.

\begin{algorithm}[h!]
	\caption{BioSequence2Vec Computation}
    \label{algo_BioSequence2Vec}
	\begin{algorithmic}[1]
 
	\State \textbf{Input:} Set ${\cal S}$ of sequences, integers $k$, $p$, $\Sigma$,$t$
	\State \textbf{Output:} Embedding $R$

    \Function{ComputeEmbedding}{$S$, $k$, $p$, $\Sigma$,$t$} 
        \State $R$ = []
        \For{$X \in S $} \Comment{for each sequence}
            \State $\vect{\hat{x}}$ = [] 
            \For{$ i = 1 \textup { to } t $} \Comment{\# of hash functions}
                \State $a_0,a_1,a_2,a_3 \gets$ \Call{random}{0, p-1}\\ 
                \Comment{Four random integers for coefficients of polynomial}
                
                \For{$ j \in \vert X \vert - k + 1 $} 
                \Comment{scan sequence}
                \State $ \alpha \gets X[j:j+k]$  \Comment{k-mer}

                    \State h $\gets$ $a_0 + a_1 \alpha_{\Sigma} + a_2 \alpha_{\Sigma}^2 + a_3 \alpha_{\Sigma}^3$\\ 
                    \Comment{$\alpha_{\Sigma}$ is numerical version of $\alpha$ base $|\Sigma|$}
                    
                    \State $h \gets (h \mod p) \mod 2$  
                    \If{$h = 0$}
                        \State $\vect{\hat{x}}[i]$ $\gets$ $\vect{\hat{x}}[i]$ - 1 \Comment{Eq.~\eqref{eq_h_val}}
                    \Else
                        \State $\vect{\hat{x}}[i]$ $\gets$ $\vect{\hat{x}}[i]$ + 1 \Comment{Eq.~\eqref{eq_h_val}}
                    \EndIf
                    
                \EndFor
                \State $\vect{\hat{x}}[i]$ = $\frac{1}{\sqrt{t}}$ $\times \; \vect{\hat{x}}[i]$ \Comment{Eq.~\eqref{sketch}}
                
            \EndFor

            \State R.append($\vect{\hat{x}}$)
        
    \EndFor
      
    \State \Return R
    \EndFunction
    
	\end{algorithmic}
\end{algorithm}

\section{Experimental Evaluation}\label{sec_datasetDes}
This section discusses datasets and state-of-the-art (SOTA) methods for comparing results. 
All experiments are performed on a core i5 system (with a $2.40$ GHz processor) having windows 10 OS and $32$ GB memory. 
For experiments, we use $70$-$30 \%$ split for training and testing (held out) sets, respectively, and repeat experiments $5$ times to report average and standard deviation (SD) results. 
To evaluate the proposed method, we use aligned and unaligned biological sequence datasets (see Table~\ref{tbl_dataset_detail}). 
For classification, we use SVM, Naive Bayes (NB), Multi-Layer Perceptron (MLP), KNN, Random Forest (RF), Logistic Regression (LR), and Decision Tree (DT). 
We use eight SOTA methods (both alignment-free and alignment-based) to compare results. The detail of SOTA methods and a brief comparison with the proposed model are given in Table~\ref{tbl_sota_detail}.

\begin{table}[h!]
    \centering
    \resizebox{1\textwidth}{!}{
    \begin{tabular}{p{1.8cm}p{5.8cm}cp{1.6cm}p{1cm}ccc}
    \toprule
      \multirow{2}{*}{Dataset} & \multirow{2}{*}{Detail} & \multirow{2}{*}{Source} & \multirow{2}{2cm}{Total Sequences} & \multirow{2}{1.5cm}{Total classes} & \multicolumn{3}{c}{Sequence Length} \\
      \cmidrule{6-8}
        & & & & &  Min & Max & Average \\
    \midrule \midrule
    \multirow{2}{2cm}{Spike7k} & Aligned spike protein sequences to classify lineages of coronavirus in humans & \multirow{2}{*}{~\cite{gisaid_website_url}} & \multirow{2}{*}{7000} & \multirow{2}{*}{22} & \multirow{2}{*}{1274} & \multirow{2}{*}{1274} & \multirow{2}{*}{1274.00} \\
    \midrule
    \multirow{2}{1.7cm}{Human DNA} & Unaligned nucleotide sequences to classify gene family to which humans belong & \multirow{2}{*}{~\cite{human_dna_website_url}} & \multirow{2}{*}{4380} & \multirow{2}{*}{7} & \multirow{2}{*}{5} & \multirow{2}{*}{18921} & \multirow{2}{*}{1263.59} \\
      \bottomrule
    \end{tabular}
    }
    \caption{Dataset Statistics.}
    \label{tbl_dataset_detail}
\end{table}

\begin{table}[h!]
    \centering
    \resizebox{1\textwidth}{!}{
    \begin{tabular}{p{2.5cm}p{2cm}p{5cm}cp{1.6cm}p{2.5cm}p{1.2cm}p{1.7cm}}
    \toprule
      Method & Category & Detail & Source & Alignment Free & Computationally Efficient & Space Efficient & Low Dim. Embedding \\
    \midrule \midrule
    Spike2Vec & \multirow{3}{2cm}{Feature Engineering} & \multirow{3}{5cm}{Take biological sequence as input and design fixed-length numerical embeddings} & ~\cite{ali2021spike2vec} & $\checkmark$ & $\checkmark$  & $\checkmark$  & $\tikzxmark$\\
    Spaced k-mers & & & ~\cite{singh2017gakco} & $\checkmark$ & $\checkmark$ & $\checkmark$  & $\tikzxmark$ \\
    PWM2Vec & & & ~\cite{ali2022pwm2vec} & $\tikzxmark$ & $\checkmark$ & $\checkmark$  & $\checkmark$ \\
    \cmidrule{2-8}
    \multirow{2}{*}{WDGRL} &  \multirow{4}{2cm}{Neural Network (NN)} &  \multirow{4}{5cm}{Take one-hot representation of biological sequence as input and design NN-based embedding method by minimizing loss} & \multirow{2}{*}{~\cite{shen2018wasserstein}} & \multirow{2}{*}{$\tikzxmark$} & \multirow{2}{*}{$\tikzxmark$}  & \multirow{2}{*}{$\checkmark$}  & \multirow{2}{*}{$\checkmark$} \\
    & \\
    \multirow{2}{*}{AutoEncoder} & & & \multirow{2}{*}{~\cite{xie2016unsupervised}} & \multirow{2}{*}{$\tikzxmark$} & \multirow{2}{*}{$\tikzxmark$} & \multirow{2}{*}{$\checkmark$}  & \multirow{2}{*}{$\checkmark$} \\
    & \\
    \cmidrule{2-8}
    \multirow{4}{*}{String Kernel} & \multirow{4}{2cm}{Kernel Matrix} & Designs $n \times n$ kernel matrix that can be used with kernel classifiers or with kernel PCA to get feature vector based on principal components & \multirow{4}{*}{~\cite{farhan2017efficient}} & \multirow{4}{*}{$\checkmark$} & \multirow{4}{*}{$\tikzxmark$} & \multirow{4}{*}{$\tikzxmark$}  & \multirow{4}{*}{$\checkmark$} \\
    \cmidrule{2-8}
    \multirow{4}{*}{SeqVec} & \multirow{4}{2cm}{Pretrained Language Model} & Takes biological sequences as input and fine-tunes the weights based on a pre-trained model to get final embedding & \multirow{4}{*}{~\cite{heinzinger2019modeling}} & \multirow{4}{*}{$\checkmark$} & \multirow{4}{*}{$\tikzxmark$} & \multirow{4}{*}{$\checkmark$}  & \multirow{4}{*}{$\checkmark$} \\
    \cmidrule{2-8}
    \multirow{4}{*}{ProteinBERT} & \multirow{4}{2cm}{Pretrained Transformer} & A pretrained protein sequence model to classify the given biological sequence using Transformer/Bert & \multirow{4}{*}{~\cite{brandes2022proteinbert}} & \multirow{4}{*}{$\checkmark$} & \multirow{4}{*}{$\tikzxmark$} & \multirow{4}{*}{$\checkmark$}  & \multirow{4}{*}{$\checkmark$} \\
    \cmidrule{2-8}
    \multirow{4}{2cm}{BioSequence2Vec (ours)} & \multirow{4}{2cm}{Hashing} & Takes biological sequence as input and design embeddings based on the kernel property of preserving pairwise distance & \multirow{4}{*}{-} & \multirow{4}{*}{$\checkmark$} & \multirow{4}{*}{$\checkmark$} & \multirow{4}{*}{$\checkmark$}  & \multirow{4}{*}{$\checkmark$} \\
      \bottomrule
    \end{tabular}
    }
    \caption{Different methods (ours and SOTA) description.}
    \label{tbl_sota_detail}
\end{table}

\section{Results and Discussion}\label{sec_results}
In this section, we report the classification results for BioSequence2Vec using different datasets and compare the results with SOTA methods.
A comparison of BioSequence2Vec with SOTA algorithms on Spike7k and the Human DNA dataset is shown in Table~\ref{tbl_results_classification}. We report the BioSequence2Vec results for $t=1000$ and $k=3$.

For the aligned Spike7k protein sequence dataset, we can observe that the proposed BioSequence2Vec with random forest classifier outperforms all the SOTA methods for all but one evaluation metric. In the case of training runtime, WDGRL performs the best because of having the smallest size embedding.

For the unaligned Human DNA (nucleotide) data, we can observe in Table~\ref{tbl_results_classification} that the random forest classifier with BioSequence2Vec outperforms all SOTA methods in all evaluation metrics except the classification training runtime. An important point to note here is that all alignment-free methods (i.e., Spike2Vec, Spaced $k$-mers, String kernel, and BioSequence2Vec) generally show better predictive performance as compared to the alignment-based methods such as PWM2Vec, WDGRL, AE. Among alignment-free methods, the proposed method performs the best (hence showing the generalizability property), showing that we can completely eliminate the multiple sequence alignment from the pipeline (an NP-hard step).
Moreover, using pre-trained language models such as SeqVec and ProteinBert also did not improve the predictive performance.

\begin{table}[h!]
\centering
\resizebox{1\textwidth}{!}{
 \begin{tabular}{@{\extracolsep{6pt}}p{2.5cm}lp{1.1cm}p{1.1cm}p{1.1cm}p{1.3cm}p{1.3cm}p{1.1cm}p{1.7cm}
 p{1.1cm}p{1.1cm}p{1.1cm}p{1.3cm}p{1.3cm}p{1.1cm}p{1.7cm}}
    \toprule
    & & \multicolumn{7}{c}{Spike7k} & \multicolumn{7}{c}{Human DNA} \\
    \cmidrule{3-9} \cmidrule{10-16}
        \multirow{2}{*}{Embeddings} & \multirow{2}{*}{Algo.} & \multirow{2}{*}{Acc. $\uparrow$} & \multirow{2}{*}{Prec. $\uparrow$} & \multirow{2}{*}{Recall $\uparrow$} & \multirow{2}{1.4cm}{F1 (Weig.) $\uparrow$} & \multirow{2}{1.5cm}{F1 (Macro) $\uparrow$} & \multirow{2}{1.2cm}{ROC AUC $\uparrow$} & Train Time (sec.) $\downarrow$
          & \multirow{2}{*}{Acc. $\uparrow$} & \multirow{2}{*}{Prec. $\uparrow$} & \multirow{2}{*}{Recall $\uparrow$} & \multirow{2}{1.4cm}{F1 (Weig.) $\uparrow$} & \multirow{2}{1.5cm}{F1 (Macro) $\uparrow$} & \multirow{2}{1.2cm}{ROC AUC $\uparrow$} & Train Time (sec.) $\downarrow$\\
        \midrule \midrule
        \multirow{7}{1.2cm}{Spike2Vec}
         & SVM & 0.855 & 0.853 & 0.855 & 0.843 & 0.689 & 0.843 & 61.112  & 0.597 & 0.602 & 0.597 & 0.589 & 0.563 & 0.733 & 4.612 \\
         & NB & 0.476 & 0.716 & 0.476 & 0.535 & 0.459 & 0.726 & 13.292  & 0.175 & 0.143 & 0.175 & 0.106 & 0.128 & 0.532 & 0.039 \\
         & MLP & 0.803 & 0.803 & 0.803 & 0.797 & 0.596 & 0.797 & 127.066  & 0.618 & 0.618 & 0.618 & 0.613 & 0.573 & 0.747 & 22.292 \\
         & KNN & 0.812 & 0.815 & 0.812 & 0.805 & 0.608 & 0.794 & 15.970  & 0.640 & 0.653 & 0.640 & 0.642 & 0.608 & 0.772 & 0.561 \\
         & RF & 0.856 & 0.854 & 0.856 & 0.844 & 0.683 & 0.839 & 21.141  & 0.752 & 0.773 & 0.752 & 0.749 & 0.736 & 0.824 & 2.558 \\
         & LR & 0.859 & 0.852 & 0.859 & 0.844 & 0.690 & 0.842 & 64.027  & 0.569 & 0.570 & 0.569 & 0.555 & 0.525 & 0.710 & 2.074 \\
         & DT & 0.849 & 0.849 & 0.849 & 0.839 & 0.677 & 0.837 & 4.286  & 0.621 & 0.624 & 0.621 & 0.621 & 0.594 & 0.765 & 0.275 \\
        \cmidrule{2-9} \cmidrule{10-16}
        \multirow{7}{1.2cm}{PWM2Vec}
        & SVM & 0.818 & 0.820 & 0.818 & 0.810 & 0.606 & 0.807 & 22.710  & 0.302 & 0.241 & 0.302 & 0.165 & 0.091 & 0.505 & 10011.3 \\
 & NB & 0.610 & 0.667 & 0.610 & 0.607 & 0.218 & 0.631 & 1.456  & 0.084 & 0.442 & 0.084 & 0.063 & 0.066 & 0.511 & 4.565 \\
 & MLP & 0.812 & 0.792 & 0.812 & 0.794 & 0.530 & 0.770 & 35.197  & 0.310 & 0.350 & 0.310 & 0.175 & 0.107 & 0.510 & 320.555 \\
 & KNN & 0.767 & 0.790 & 0.767 & 0.760 & 0.565 & 0.773 & 1.033  & 0.121 & 0.337 & 0.121 & 0.093 & 0.077 & 0.509 & 2.193 \\
 & RF & 0.824 & 0.843 & 0.824 & 0.813 & 0.616 & 0.803 & 8.290  & 0.309 & 0.332 & 0.309 & 0.181 & 0.110 & 0.510 & 65.250 \\
 & LR & 0.822 & 0.813 & 0.822 & 0.811 & 0.605 & 0.802 & 471.659  & 0.304 & 0.257 & 0.304 & 0.167 & 0.094 & 0.506 & 23.651 \\
 & DT & 0.803 & 0.800 & 0.803 & 0.795 & 0.581 & 0.791 & 4.100  & 0.306 & 0.284 & 0.306 & 0.181 & 0.111 & 0.509 & 1.861 \\
         \cmidrule{2-9} \cmidrule{10-16}
        \multirow{7}{1.9cm}{String Kernel}
        & SVM  & 0.845 & 0.833 & 0.846 & 0.821 & 0.631 & 0.812 & 7.350  & 0.618 & 0.617 & 0.618 & 0.613 & 0.588 & 0.753 & 39.791 \\
         & NB   & 0.753 & 0.821 & 0.755 & 0.774 & 0.602 & 0.825 & 0.178  & 0.338 & 0.452 & 0.338 & 0.347 & 0.333 & 0.617 & 0.276 \\
         & MLP  & 0.831 & 0.829 & 0.838 & 0.823 & 0.624 & 0.818 & 12.652  & 0.597 & 0.595 & 0.597 & 0.593 & 0.549 & 0.737 & 331.068 \\
         & KNN  & 0.829 & 0.822 & 0.827 & 0.827 & 0.623 & 0.791 & 0.326  & 0.645 & 0.657 & 0.645 & 0.646 & 0.612 & 0.774 & 1.274 \\
         & RF   & 0.847 & 0.844 & 0.841 & 0.835 & 0.666 & 0.824 & 1.464  & 0.731 & 0.776 & 0.731 & 0.729 & 0.723 & 0.808 & 12.673 \\
         & LR   & 0.845 & 0.843 & 0.843 & 0.826 & 0.628 & 0.812 & 1.869  & 0.571 & 0.570 & 0.571 & 0.558 & 0.532 & 0.716 & 2.995 \\
         & DT   & 0.822 & 0.829 & 0.824 & 0.829 & 0.631 & 0.826 & 0.243  & 0.630 & 0.631 & 0.630 & 0.630 & 0.598 & 0.767 & 2.682 \\
          \cmidrule{2-9} \cmidrule{10-16}
           \multirow{7}{1.2cm}{WDGRL}  & SVM & 0.792 & 0.769 & 0.792 & 0.772 & 0.455 & 0.736 & 0.335 & 0.318 & 0.101 & 0.318 & 0.154 & 0.069 & 0.500 & 0.751 \\
             & NB & 0.724 & 0.755 & 0.724 & 0.726 & 0.434 & 0.727 & 0.018  & 0.232 & 0.214 & 0.232 & 0.196 & 0.138 & 0.517 & \textbf{0.004} \\
             & MLP & 0.799 & 0.779 & 0.799 & 0.784 & 0.505 & 0.755 & 7.348  & 0.326 & 0.286 & 0.326 & 0.263 & 0.186 & 0.535 & 8.613 \\
             & KNN & 0.800 & 0.799 & 0.800 & 0.792 & 0.546 & 0.766 & 0.094  & 0.317 & 0.317 & 0.317 & 0.315 & 0.266 & 0.574 & 0.092 \\
             & RF & 0.796 & 0.793 & 0.796 & 0.789 & 0.560 & 0.776 & 0.393  & 0.453 & 0.501 & 0.453 & 0.430 & 0.389 & 0.625 & 1.124 \\
             & LR & 0.752 & 0.693 & 0.752 & 0.716 & 0.262 & 0.648 & 0.091  & 0.323 & 0.279 & 0.323 & 0.177 & 0.095 & 0.507 & 0.041 \\
             & DT & 0.790 & 0.799 & 0.790 & 0.788 & 0.557 & 0.768 & \textbf{0.009}  & 0.368 & 0.372 & 0.368 & 0.369 & 0.328 & 0.610 & 0.047 \\
 \cmidrule{2-9} \cmidrule{10-16}
 \multirow{7}{1.9cm}{Spaced $k$-mers} 
            & SVM & 0.852 & 0.841 & 0.852 & 0.836 & 0.678 & 0.840 & 2218.347  & 0.746 & 0.749 & 0.746 & 0.746 & 0.728 & 0.844 & 26.957 \\
            & NB & 0.655 & 0.742 & 0.655 & 0.658 & 0.481 & 0.749 & 267.243  & 0.177 & 0.233 & 0.177 & 0.122 & 0.142 & 0.533 & 0.467  \\ 
            & MLP & 0.809 & 0.810 & 0.809 & 0.802 & 0.608 & 0.812 & 2072.029  & 0.722 & 0.723 & 0.722 & 0.720 & 0.689 & 0.817 & 126.584  \\
            & KNN & 0.821 & 0.810 & 0.821 & 0.805 & 0.591 & 0.788 & 55.140  & 0.699 & 0.704 & 0.699 & 0.698 & 0.667 & 0.804 & 1.407  \\
            & RF & 0.851 & 0.842 & 0.851 & 0.834 & 0.665 & 0.833 & 646.557  & 0.784 & 0.814 & 0.784 & 0.782 & 0.773 & 0.843 & 13.397  \\
            & LR & 0.855 & 0.848 & 0.855 & 0.840 & 0.682 & 0.840 & 200.477  & 0.712 & 0.712 & 0.712 & 0.709 & 0.693 & 0.812 & 37.756  \\
            & DT & 0.853 & 0.850 & 0.853 & 0.841 & 0.685 & 0.842 & 98.089  & 0.656 & 0.658 & 0.656 & 0.656 & 0.626 & 0.784 & 2.985  \\
  \cmidrule{2-9} \cmidrule{10-16}
\multirow{7}{1.5cm}{Auto-Encoder}
 & SVM &  0.699 & 0.720 & 0.699 & 0.678 & 0.243 & 0.627 & 4018.028  & 0.621 & 0.638 & 0.621 & 0.624 & 0.593 & 0.769 & 22.230 \\
 & NB & 0.490 & 0.533 & 0.490 & 0.481 & 0.123 & 0.620 & 24.6372  & 0.260 & 0.426 & 0.260 & 0.247 & 0.268 & 0.583 & 0.287 \\
 & MLP & 0.663 & 0.633 & 0.663 & 0.632 & 0.161 & 0.589 & 87.4913  & 0.621 & 0.624 & 0.621 & 0.620 & 0.578 & 0.756 & 111.809 \\
 & KNN & 0.782 & 0.791 & 0.782 &  0.776 & 0.535 & 0.761 & 24.5597  & 0.565 & 0.577 & 0.565 & 0.568 & 0.547 & 0.732 & 1.208 \\
 & RF & 0.814 & 0.803 & 0.814 & 0.802 & 0.593 & 0.793 &  46.583  & 0.689 & 0.738 & 0.689 & 0.683 & 0.668 & 0.774 & 20.131 \\
 & LR & 0.761 & 0.755 & 0.761 & 0.735 & 0.408 & 0.705 & 11769.02  & 0.692 & 0.700 & 0.692 & 0.693 & 0.672 & 0.799 & 58.369 \\
 & DT & 0.803 & 0.792 & 0.803 & 0.792 & 0.546 & 0.779 & 102.185  & 0.543 & 0.546 & 0.543 & 0.543 & 0.515 & 0.718 & 10.616 \\
  \cmidrule{2-9} \cmidrule{10-16}
\multirow{7}{1.5cm}{SeqVec}
 & SVM & 0.796 & 0.768 & 0.796 & 0.770 & 0.479 & 0.747 & 1.0996  & 0.656 & 0.661 & 0.656 & 0.652 & 0.611 & 0.791 & 0.891 \\
 & NB & 0.686 & 0.703 & 0.686 & 0.686 & 0.351 & 0.694 & 0.0146  & 0.324 & 0.445 & 0.312 & 0.295 & 0.282 & 0.624 & 0.036 \\
 & MLP & 0.796 & 0.771 & 0.796 & 0.771 & 0.510 & 0.762 & 13.172  & 0.657 & 0.633 & 0.653 & 0.646 & 0.616 & 0.783 & 12.432 \\
 & KNN & 0.790 & 0.787 & 0.790 & 0.786 & 0.561 & 0.768 &  0.6463  & 0.592 & 0.606 & 0.592 & 0.591 & 0.552 & 0.717 & 0.571 \\
 & RF & 0.793 & 0.788 & 0.793 & 0.786 & 0.557 & 0.769 &  1.8241  & 0.713 & 0.724 & 0.701 & 0.702 & 0.693 & 0.752 & 2.164 \\
 & LR & 0.785 & 0.763 & 0.785 & 0.761 & 0.459 & 0.740 & 1.7535  & 0.725 & 0.715 & 0.726 & 0.725 & 0.685 & 0.784 & 1.209 \\
 & DT & 0.757 & 0.756 & 0.757 & 0.755 & 0.521 & 0.760 & 0.1308   & 0.586 & 0.553 & 0.585 & 0.577 & 0.557 & 0.736 & 0.24 \\
 \cmidrule{2-9} \cmidrule{10-16}
\multirow{1}{1.5cm}{Protein Bert}
 & \_ &  0.836 & 0.828 & 0.836 & 0.814 & 0.570 & 0.792 & 14163.52  &  0.542 & 0.580 & 0.542 & 0.514 & 0.447 & 0.675 & 58681.57 \\
 \cmidrule{2-9} \cmidrule{10-16}
\multirow{7}{1.5cm}{BioSequence2Vec (ours)}
 & SVM & 0.848 & 0.858 & 0.848 & 0.841 & 0.681 & 0.848 & 9.801  & 0.555 & 0.554 & 0.555 & 0.543 & 0.497 & 0.700 & 13.251 \\
 & NB & 0.732 & 0.776 & 0.732 & 0.741 & 0.555 & 0.771 & 1.440  & 0.263 & 0.518 & 0.263 & 0.244 & 0.239 & 0.572 & 0.095 \\
 & MLP & 0.835 & 0.825 & 0.835 & 0.825 & 0.622 & 0.819 & 13.893  & 0.583 & 0.598 & 0.583 & 0.571 & 0.541 & 0.717 & 70.463 \\
 & KNN & 0.821 & 0.818 & 0.821 & 0.811 & 0.616 & 0.803 & 1.472  & 0.613 & 0.625 & 0.613 & 0.615 & 0.565 & 0.748 & 0.313 \\
 & RF & \textbf{0.863} & \textbf{0.867} & \textbf{0.863} & \textbf{0.854} & \textbf{0.703} & \textbf{0.851} & 2.627  & \textbf{0.786} & \textbf{0.816} & \textbf{0.786} & \textbf{0.787} & \textbf{0.779} & \textbf{0.846} & 1.544 \\
 & LR & 0.500 & 0.264 & 0.500 & 0.333 & 0.031 & 0.500 & 11.907  & 0.527 & 0.522 & 0.527 & 0.501 & 0.457 & 0.674 & 29.029 \\
 & DT & 0.845 & 0.856 & 0.845 & 0.841 & 0.683 & 0.839 & 0.956  & 0.663 & 0.666 & 0.663 & 0.664 & 0.639 & 0.795 & 4.064 \\
         \bottomrule
         \end{tabular}
}
 \caption{Classification results (averaged over $5$ runs) on \textbf{Spike7k} and \textbf{Human DNA} datasets for different evaluation metrics. Best values are shown in bold.}
    \label{tbl_results_classification}
\end{table}

For ProteinBert, the main reason for its comparable performance to BioSequence2Vec on Spike7k data while bad performance on Human DNA data is because it is pretrained on protein sequences in the original study, hence performing badly on Human DNA Nucleotide data (poor generalizability). Although SeqVec is also pretrained on protein sequences (in the original study), its comparatively better performance on nucleotide data is because we use it to design the embeddings and then use ML classifiers for the prediction, which performs better for tabular data compared to DL models~\cite{shwartz2022tabular}.
To check if the computed results are statistically significant, we used the student t-test and observed the $p$-values using average and standard deviations (SD) of $5$ runs. 
We noted that $p$-values were $< 0.05$ in the majority of the cases (because SD values are very low), confirming the statistical significance of the results.

\section{Conclusion}\label{sec_conclusion}
In this paper, we propose an efficient and alignment-free method, called BioSequence2Vec, to generate embeddings for biological sequences using the idea of hashing. We show that BioSequence2Vec has the qualities of kernel methods while being
memory efficient. We performed extensive experiments on real-world biological sequence data to validate the proposed model using different evaluation metrics. BioSequence2Vec outperforms the SOTA models in terms of predictive accuracy.  Future work involves
evaluating BioSequence2Vec on millions of sequences and other virus data.  Applying this method to
other domains (\eg, music or video) would also be
an interesting future extension.

\bibliographystyle{splncs04}
\bibliography{references}

\end{document}